\documentclass[runningheads]{llncs}
\usepackage{graphicx}
\usepackage{booktabs}
\usepackage[colorlinks,linkcolor=blue]{hyperref}
\usepackage[table,xcdraw]{xcolor}
\usepackage[misc]{ifsym} 
\begin{document}
\title{Masked Video Modeling with Correlation-aware Contrastive Learning for Breast Cancer Diagnosis in Ultrasound}
\author{
Zehui Lin\inst{1,2,3}\and
Ruobing Huang\inst{1,2,3}\textsuperscript{(\Letter)}\and
Dong Ni\inst{1,2,3}\and
Jiayi Wu\inst{4} \and
Baoming Luo\inst{4}
}

\authorrunning{Z. Lin et al.}
\titlerunning{MVCC Framework for Breast Ultrasound Video Diagnosis}
\institute{
  \textsuperscript{$1$} National-Regional Key Technology Engineering Laboratory for Medical Ultrasound, School of Biomedical Engineering, Health Science Center, Shenzhen University, China\\
  \email{ruobing.huang@szu.edu.cn} \\
  \textsuperscript{$2$} Medical Ultrasound Image Computing (MUSIC) Lab, Shenzhen University, China\\
  \textsuperscript{$3$} Marshall Laboratory of Biomedical Engineering, Shenzhen University, China\\
  \textsuperscript{$4$} Department of Ultrasound, Sun Yat-Sen Memorial Hospital of Sun Yat-Sen University, China\\}
\maketitle

\begin{abstract}
  Breast cancer is one of the leading causes of cancer deaths in women. As the primary output of breast screening, breast ultrasound (US) video contains exclusive dynamic information for cancer diagnosis. However, training models for video analysis is non-trivial as it requires a voluminous dataset which is also expensive to annotate. Furthermore, the diagnosis of breast lesion faces unique challenges such as inter-class similarity and intra-class variation. In this paper, we propose a pioneering approach that directly utilizes US videos in computer-aided breast cancer diagnosis. It leverages masked video modeling as pretraining to reduce reliance on dataset size and detailed annotations. Moreover, a correlation-aware contrastive loss is developed to facilitate the identifying of the internal and external relationship between benign and malignant lesions. Experimental results show that our proposed approach achieved promising classification performance and can outperform other state-of-the-art methods.
\end{abstract}

\section{Introduction}
\begin{figure}
    \centering
    \includegraphics[width=0.8\textwidth]{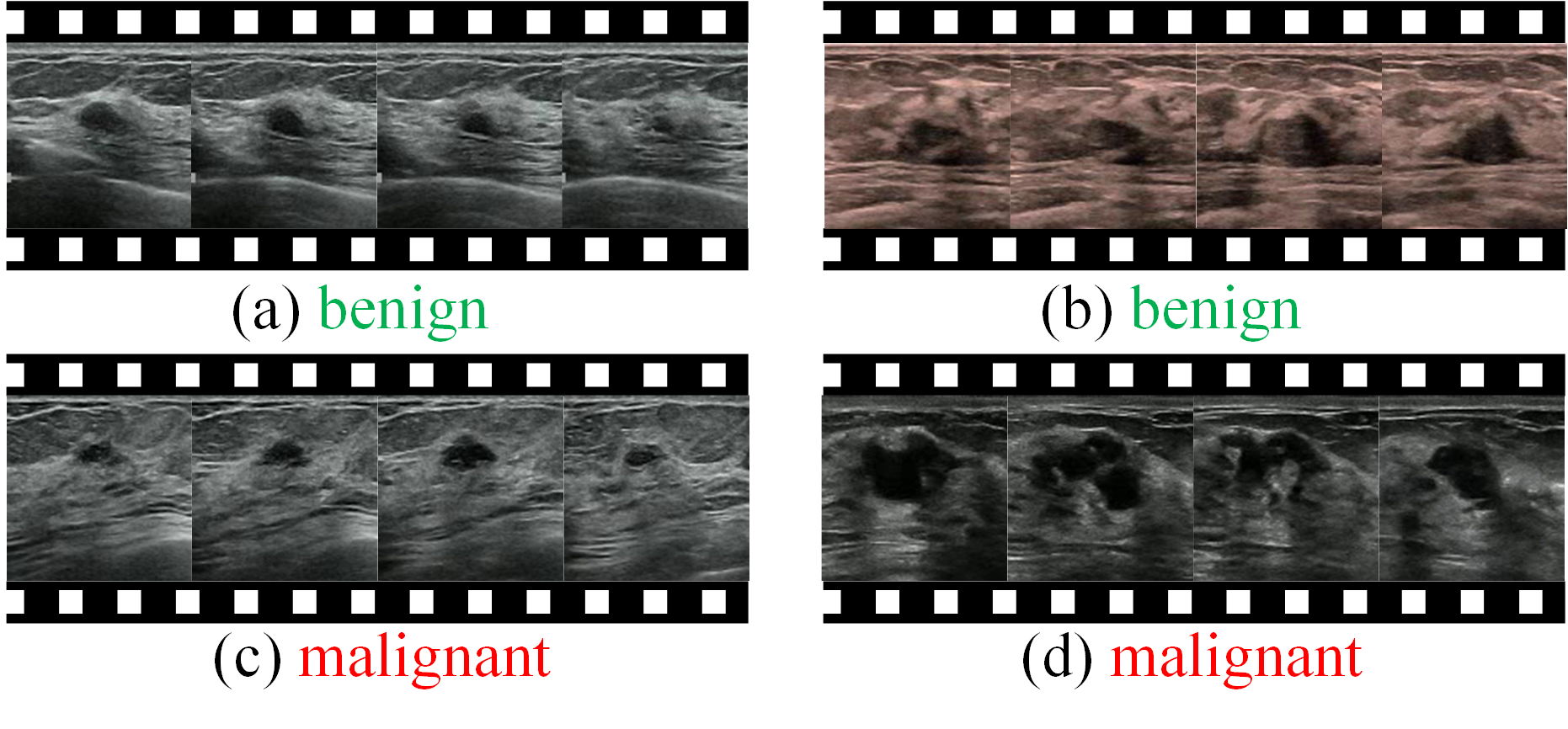}
    \caption{
        Breast ultrasound video examples. (a), (b) indicate benign cases and (c), (d) indicate malignant ones. A video contains hundreds of frames, each of which comprises a full 2D US image. Each video has a binary label (i.e. benign or malignant)--extreme sparse supervision signals compare to the data size. Besides, the lesions have a diverse appearance and complex surrounding tissues that brings additional challenges to the classification.
    }
    \label{fig:intro}
\end{figure}

Being painless, cost-effective and radiation-free, ultrasound (US) imaging is widely used in breast cancer screening \cite{brem2015screening}, especially for the evaluation of dense breasts \cite{nothacker2009early}. Its real-time imaging capability allows rapid acquisition of tissue information and produces corresponding US image sequences (i.e., videos) on the screen instantaneously. Compared with the single conventional 2D static image, the original US video of a lesion contains richer spatial-temporal information and is beneficial for diagnosis \cite{yang2021section,youk2016comparison}. However, deciphering raw US videos faces several challenges, including heavy data-oriented dependency, scarce supervision signals, inter-class similarity and intra-class variance (see Fig.~\ref{fig:intro}). A tailored computer-aided diagnostic (CAD) tool is needed to address these challenges and better assist clinicians to prevent misdiagnosis and overtreatment.

A considerable amount of 2D static image works have been proposed to help sonographers in diagnosing breast cancer \cite{chan2019cad}. Traditional methods typically extract handcraft textural features. For example, Flores et al. \cite{flores2015improving} analyzed morphological and texture features and used a features selection methodology to improve the classification performance of breast tumors on ultrasonography. Recently, deep learning models are favoured in designing new CAD tools due to their strong representation learning ability. For example, Zeimarani et al. \cite{zeimarani2020breast} used a custom-built convolutional neural network (CNN) with a few hidden layers and applied regularization techniques to improve the diagnostic performance using 2D US. These 2D approaches have demonstrated promising results while they are unable to handle video data with excessive dimension and scarce labels. Some researchers have investigated ways to directly utilize US videos of other organs or other modalities. In \cite{chen2021uscl}, Chen et al. exploited contrastive learning to initialize video model for US videos of lung and liver. Another group focused on contrast-enhanced US \cite{chen2021domain} and proposed to add additional attention modules on a 3D CNN backbone to classify lesions. These methods have their own merits while they neglect the natural spatiotemporal patterns that lie within the video. 

In this paper, we propose a novel video classification framework, named Masked Video modeling with Correlation-aware Contrastive learning (MVCC), to address the challenges in breast cancer diagnosis. Our contribution is three-fold. First, to the best of our knowledge, this is the first study that directly utilizes the videos of common B-mode Breast US for breast nodule identification. Second, we proposed to use masked video modeling to fully exploit limited data and exiguous annotations. Different from the 2D approach~\cite{he2021masked}, a dual-level masking strategy is adopted to explicitly extract features in both the spatial and temporal dimensions. Third, we selectively constrain the high-level representations to combat intra-class variation and inter-class similarity through a novel correlation-aware contrastive loss. 
Validation experiments showed that the proposed method was able to process rich video information and identify the breast nodules accurately.

\section{Methodology}
\begin{figure}
    \centering
    \includegraphics[width=\textwidth]{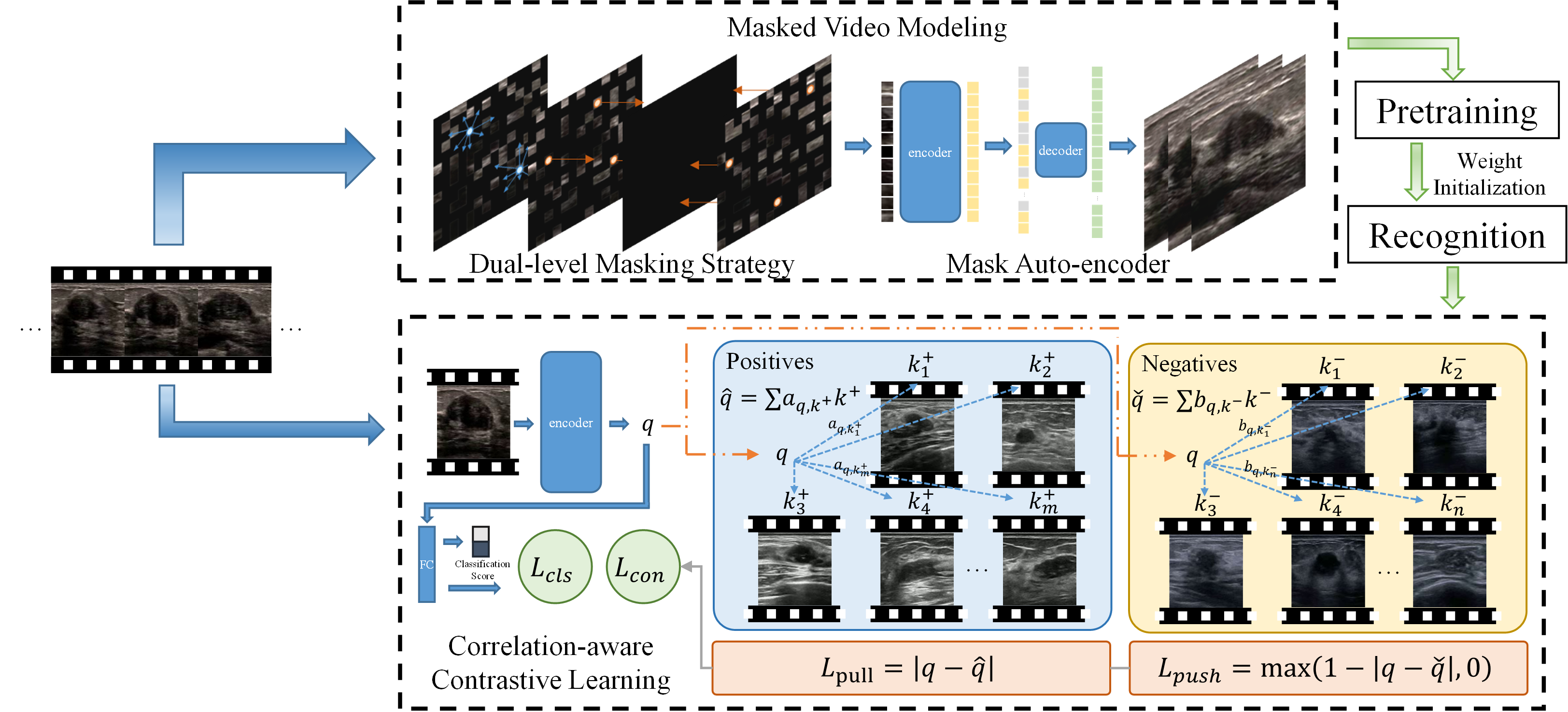}
    \caption{
        Overview of our proposed framework.
    }
    \label{framework}
\end{figure}

To fully exploit the limited annotation and available data, we propose to equip the classification model with masked video modeling and correlation-aware contrastive learning. Fig.~\ref{framework} displays the overall framework. A TimeSfomer-based \cite{bertasius2021space} auto-encoder is first built and pre-trained with a novel dual-level masking strategy to fully capture the spatial and temporal dependencies in a self-supervised manner. The weight of the trained encoder is then retained and fine-tuned using the ground truth labels. Furthermore, the model is constrained by a correlation-aware contrastive loss to selectively encourage feature resemblance among the same class, while explicitly penalizing this among different classes. Details are explained in the following section. 

\subsection{Masked Video Modeling}
In video analysis, state-of-the-art deep learning models with growing capacity and capability can easily overfit, even on large datasets. Meanwhile, annotating video dataset is extremely time- and labour-demanding. As a result, training deep learning based video analysis model is non-trivial. On the other hand, videos naturally contain rich spatial-temporal context information, reflecting the structures of normal tissues and lesions. To exploit this, we leverage self-supervised learning to prepare the model for the downstream video classification task. In specific, we opt for an auto-encoder approach that reconstructs the original video given its partial observation. In other words, a video $x$ is masked based on strategy $S: x \rightarrow \hat{x}$. $\hat{x}$ is then passed to an auto-encoder $De(En(\hat{x}))$ to reconstruct the original video from the masked data by minimizing the difference between its output $\overline{x}$ and $x$, $s.t.~\overline{x}=De(En(\hat{x}))$. $En()$ represents the encoder part while $De()$ represents the decoder. This pre-training task helps to learn informative low-level patterns and understand the global context contains in videos, consequently, could provide a good initialization for the subsequent downstream task.

\noindent
\textbf{Dual-level Masking Strategy.} Some studies have investigated image-based masking strategies, while extending this technique to video faces the additional temporal dimension and more severe information redundancy. To explicitly extract information from both spatial and temporal dimensions,  we propose a dual-level masking strategy. In particular, define an input video $x=\{f_a\}_1^t, a=1,2,...t$. The $t$ frames of $x$ is randomly selected and masked, producing $x'=\{f_{a'}\}_1^{t'}, t'<t$. The frame-level masking ratio is controlled by $\alpha$, which satisfies $t' = t\times{\alpha}$. Next, each frame $f_{a'}$ in $x'$ are divided into regular non-overlapping patches for patch-level masking. $\beta \%$ of patches are removed and obtain the masked frame $f'_{a'}$. 
Finally, we obtain the masked video $\hat{x}=\{f'_{a'}\}_1^{t'}$. The whole dual-level masking strategy can be summarized as $S: x \rightarrow x' \rightarrow \hat{x}$, and $\alpha + (1-\alpha)\beta\%$ signals are masked in total. This dual-level masking strategy explicitly forces the removal of both spatial and temporal information, creating a more suitable pre-training task for video analysis. As can be seen in Fig.~\ref{framework}, the patch-level masking (blue dots and arrows) erases local patches and requires comprehension over the tissue appearance for restoration. The frame-level masking, on the other hand, breaks connectivity in the temporal space (orange dots and arrows) and can only be recovered based on understanding over global anatomical configuration. Later experimental results also demonstrate the efficacy of this dual-level design for exploiting video data. 

\noindent
\textbf{Masked Auto-Encoder.} Our encoder leverages the TimeSformer \cite{bertasius2021space} backbone which applies temporal and spatial attention separately in sub-blocks. Inspired by \cite{he2021masked}, we design a decoder using a lightweight transformer structure \cite{he2021masked} and is only used during the pre-training phase. The input of the MAE decoder is the full set of tokens consisting of (i) encoded visible signal, and (ii) masked tokens. Positional embeddings \cite{dosovitskiy2020image} are added to all tokens to provide location information. The decoder reconstructs the input pixel values for each masked token and we use mean squared error (MSE) loss to compute the difference between $\hat{x}$ and $x$ in the pixel space (only on masked patches). The asymmetric design over encoder and decoder creates an opportunity for saving computation and training time. 

\subsection{Correlation-aware Contrastive Learning}
The core of contrastive learning is to learn the representations which maximize the agreement between the similar, related instances (i.e., positives) and minimize the similarity between the different and unrelated instances (i.e., negatives) at the same time. Supervised contrastive learning~\cite{khosla2020supervised} leverages the label information to construct the positive and negative samples and have shown promising results on many applications. However, it overlooks the fact that samples from the same class could have a dramatically different appearance (see Fig.~\ref{fig:intro} (c) and (d)). Imposing similarity constraints to these samples might confuse the model and jeopardize convergence. Furthermore, the most challenging examples to classify are those who share similar features while belonging to different classes (see Fig.~\ref{fig:intro} (a) and (c)). Therefore, we propose to take the correlation among samples into account in contrastive learning. In other words, correlated, ambiguous negative samples are heavily penalized to maximize the margin between decision surface. Meanwhile, the impact of uncorrelated positive samples is de-emphasized to allow flexible representation.

Formally, a query video $x$ is passed to the encoder to produce high-level representation $q=En(x)$. Within a training batch, the representation of positive samples (sharing same label with the query) can be denoted as $k_i^+ \in P = \{k_1^+, k_2^+, \cdots, k_m^+\}$ and the negative samples (sharing different labels with the query) is denoted as $k_i^- \in N = \{k_1^-, k_2^-, \cdots, k_n^-\}$. The positively-correlated representation $\hat{q}$ is calculated as: 
\begin{equation}
    \hat{q} = \sum_{k_i^+\in{P}}{a_{q,k_i^+}k_i^+},
\end{equation}
where $a_{q,k_i^+}$ is the normalized similarity of $q$ and $k_i^+$, which is defined as:
\begin{equation}
    a_{q,k_i^+} = \frac{exp\left(sim\left(q,k_i^+\right)/\tau\right)}{\sum_{k^{'}\in{P}}exp\left(sim\left(q,k^{'}\right)/\tau\right)},
\end{equation}
where $\tau$ is the temperature and $sim(y,z) = y^{T}z/\Vert{y}\Vert\Vert{z}\Vert$ measures the cosine similarity. The negatively-correlated representation $\check{q}$ is calculated in the same manner as:
\begin{equation}
    \check{q} = \sum_{k_i^-\in{N}}{b_{q,k_i^-}k_i^-},
\end{equation}
where $b_{q,k_i^-}$ is defined as: 
\begin{equation}
    b_{q,k_i^-} = \frac{exp\left(sim\left(q,k_i^-\right)/\tau\right)}{\sum_{k^{'}\in{N}}exp\left(sim\left(q,k^{'}\right)/\tau\right)}.
\end{equation}
The core of this procedure is to reconstruct the query sample based on its proximity with each positive (or negative) sample in high-level semantics. This design can highlight similar negative samples while understating the dis-similar positive ones, producing smoothed and correlation-sensible representations to calculate the contrastive loss. To combat the varying issues in differentiating the query sample against different classes, we compose the loss function with two elements. The positive element utilizes $\hat{q}$, and can be defined as:

\begin{equation}
    L_{pull} = \vert q - \hat{q} \vert.
    \label{loss_pull}
\end{equation}
Similarly, the negative element can be defined as:
\begin{equation}
    L_{push} = max\left(1 - \vert q - \check{q} \vert, 0\right),
    \label{loss_push}
\end{equation}
forming the full correlation-aware contrastive loss as:
\begin{equation}
    L_{con} = L_{pull} + L_{push}.
\end{equation}
$\vert\cdot\vert$ represents $L_1$ loss. $max\left(\cdot, 0\right) $ indicates the Hinge loss~\cite{platt1998sequential} which is able to maintain equilibrium in the optimization. $L_{con}$, therefore, neglects dis-similar positive samples to and only pulls together the similar one. Meanwhile, it deliberately pushes away similar negative samples while neglecting the distant ones which can be easily classified. In this way, the model allows distinct positive samples to have different features to handle intra-class variation, while mining hard negative examples to combat inter-class similarity.  

The overall objective is a combination of the standard cross-entropy loss $L_{cls}$ and the correlation-aware contrastive loss $L_{con}$, defined as:
\begin{equation}
    L = L_{cls} + \lambda L_{con},
    \label{loss_total}
\end{equation}
where $\lambda$ is a balancing factor of the two learning targets.

\section{Experiments and Results}

\textbf{Dataset and Implementation Details.} The in-house dataset contains 3854 breast ultrasound videos (with size 224x224, average 260 frames) approved by the local IRB. The videos were collected by different operators using different US machines. Each video contains one lesion of a patient, while its corresponding ground truth is derived from the biopsy result. Overall, the benign-to-malignant ratio is approximately 3 to 1. The dataset was randomly split into 2697, 385 and 772 videos for training, validation and independent testing. All experiments were repeated for 4 times with different random seeds and the averaged performance was reported to produce statistically stable results. The value of $\lambda$ in Eq.~\ref{loss_total} was empirically set to 0.1. We use the SGD optimizer with a learning rate of 0.0075. Different augmentation strategies were applied, including scaling, flipping, and cropping. We implemented our method in \textit{Pytorch}, using an NVIDIA RTX 3090 GPU. In this study, model performance is evaluated using accuracy (ACC(\%)), sensitivity (SEN(\%)), specificity (SPE(\%)), precision (PRE(\%)) and F1-score(\%).

\noindent
\textbf{Quantitative Analysis.} 
We first investigate whether and how the performance of the classification model changes with or without different contrastive learning strategy (conducted without using masked video modeling). We select the state-of-the-art video recognition model--TimeSformer \cite{bertasius2021space} as a strong baseline (trained only using cross-entropy). We then apply the classical supervised contrastive loss~\cite{khosla2020supervised} (denoted as `+SCL') and the proposed correlation-aware contrastive loss (denoted as `+C-SCL').

Experimental results are reported in row 1-3
Table.~\ref{tab:1}. The TimeSformer baseline alone obtained good accuracy (89.15\%), while the sensitivity is relatively low (73.84\%). This may result from fewer true positive predictions. The SCL helped to further increase the accuracy and specificity (row 2, Table.~\ref{tab:1}), which may result from its ability to constraint representations in the embedding space. However,  its sensitivity score is even lower (73.54\%), indicating a strong bias to the dominant negative class. On the other hand, the proposed C-SCL increases the sensitivity by 3.02\%, which is critical for identifying malignant nodules. This may stem from that the C-SCL explicitly pull away representations of malignant cases from similar benign ones. Furthermore, it avoids penalizing dissimilar features from the same class and may help to alleviate over-fitting. As a result, the Baseline+C-SCL combination produced the highest F1-score (80.59\%). Also, note that the design of the C-SCL is general and does not require alteration of the model architecture. It could be easily extended to other image analysis tasks without bell and whistle.

\begin{table}[htbp]
    \centering
    \caption{The mean (std) results of different methods on the breast video classification.}
    \label{tab:1}
    \begin{tabular}{@{}c|c|c|c|c|c@{}}
    \toprule
    {\color[HTML]{000000} \textbf{Method}} & \textbf{ACC(\%)} & \textbf{SEN(\%)} & \textbf{SPE(\%)} & \textbf{PRE(\%)} & \textbf{F1-score(\%)} \\ \midrule
    Baseline                               & 89.15(0.65)            & 73.84(1.34)            & 94.71(1.25)            & 83.68(3.09)            & 78.38(1.11)                 \\
    +SCL~\cite{khosla2020supervised}                                    & 89.99(1.30)            & 73.54(5.23)            & \textcolor[rgb]{0.00,0.00,1.00}{95.94(0.34)}            & \textcolor[rgb]{0.00,0.00,1.00}{86.75(1.20)}            & 79.52(3.55)                 \\
    +C-SCL                                  & \textcolor[rgb]{0.00,0.00,1.00}{90.28(1.06)}            & \textcolor[rgb]{0.00,0.00,1.00}{76.56(6.32)}            & 95.23(0.83)            & 85.46(1.29)            & \textcolor[rgb]{0.00,0.00,1.00}{80.59(3.32)}                 \\
    \midrule
    USCL~\cite{chen2021uscl}          & 83.72(1.08)            & 51.27(6.25)            & 94.52(1.53)            & 76.12(3.82)            & 60.98(4.71)                 \\
    CVRL~\cite{qian2021spatiotemporal}          & 79.21(0.19)            & 43.89(10.99)            & 92.07(3.89)            & 68.38(5.64)            & 52.14(5.80)                 \\
    MVCC-Patch          & 90.42(1.50)            & 74.16(3.84)            & \textcolor[rgb]{0.00,0.00,1.00}{96.30(0.98)}            & \textcolor[rgb]{0.00,0.00,1.00}{87.85(3.36)}            & 80.41(3.50)                 \\
    MVCC-Frame           & 90.28(0.95)            & 75.54(1.90)            & 95.64(1.19)            & 86.34(3.54)            & 80.54(1.97)                 \\
    MVCC     & \textcolor[rgb]{0.00,0.00,1.00}{90.64(1.10)}            & \textcolor[rgb]{0.00,0.00,1.00}{76.02(1.83)}            & 95.95(1.07)            & 87.21(3.27)            & \textcolor[rgb]{0.00,0.00,1.00}{81.21(2.27)}                 \\ \bottomrule
    
    \end{tabular}
\end{table}

Next, we discuss whether masked video modeling can provide a better initialization and the influence of different masking strategies. In specific, two state-of-the-art (SOTA) video pretraining methods: USCL~\cite{chen2021uscl} and CVRL~\cite{qian2021spatiotemporal} are compared with our method. We use the patch-level only, frame-level only and the proposed dual-level masking strategy respectively to pre-train the classification model. Note that the patch-level strategy is essentially similar to~\cite{he2021masked}, which was proposed to analyse 2D images. 

Table.~\ref{tab:1} shows the results (row 4-8). As we can see that the SOTA methods gain low sensitivity and F1-score (row 4-5). We conjecture this may be due to some hard examples sharing similar features while belonging to different classes failing these methods, resulting in poor initialization and local optimum. Compared with natural video dataset k400 \cite{kay2017kinetics} initialization (row 3) the patch-level masking approach obtained higher accuracy, specificity and precision, proves itself as a useful initialization technique to the normal natural data pretraining. It is interesting to see the frame-level masking strategy alone does not lead to performance gain (compare row 7 to row 3). This may result from that there exists information redundancy and the frame-level masking cannot assist the extraction of anatomical features than lies within the spatial dimension. On the contrary, the proposed dual-level masking strategy utilizes both spatial and temporal information and scored the highest accuracy (90.64\%) and F1-score (81.21\%) among all competing methods. It is worth noting that despite facing an imbalanced dataset (fewer malignant samples), our approach were still able to gain high sensitivity, and ultimately the overall performance. 

\begin{figure}[htbp]
    \centering
    \includegraphics[width=0.7\textwidth]{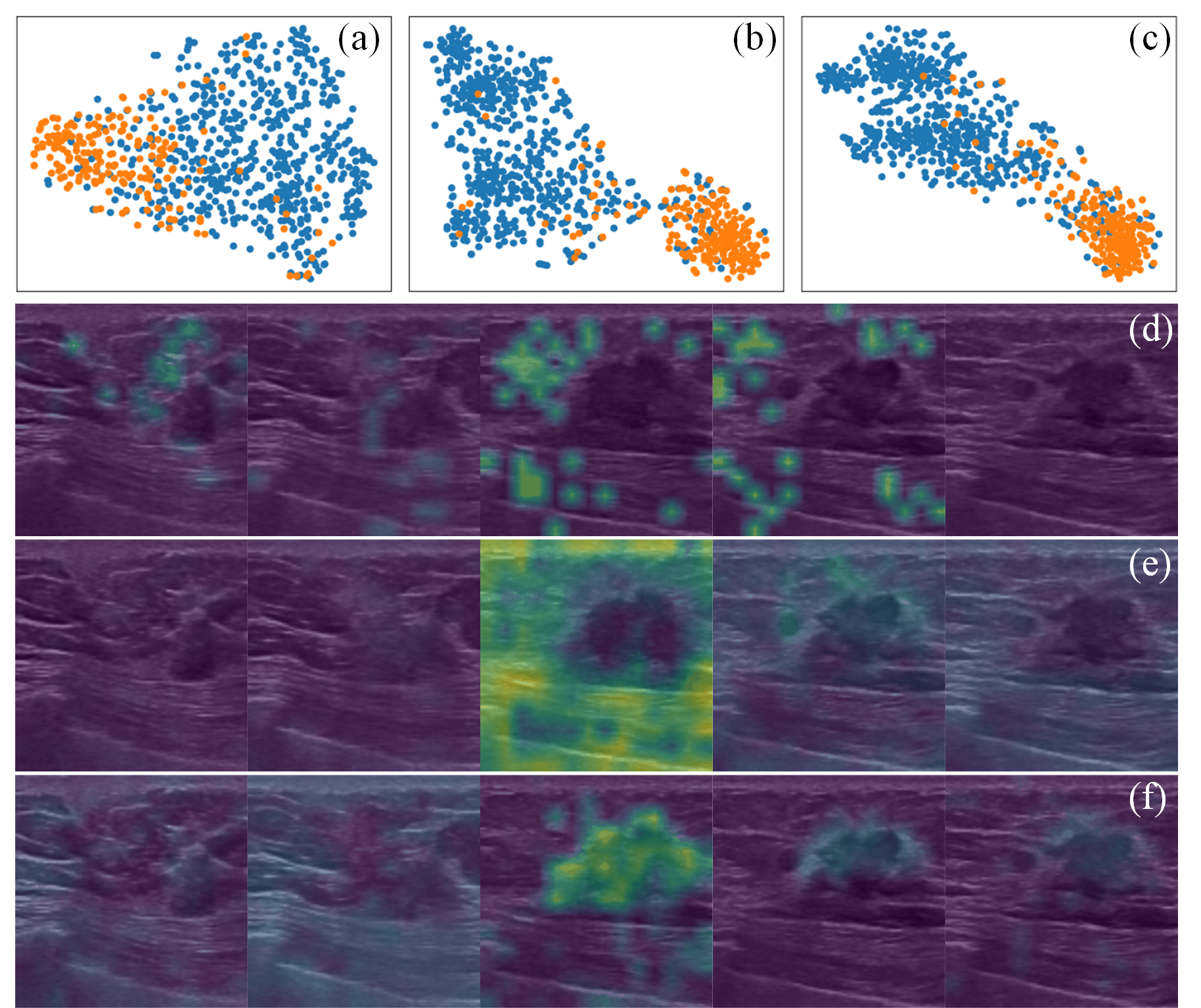}
    \caption{
        T-SNE and Grad-CAM visualization. (a) and (d) correspond to the baseline. (b) and (e) correspond to the correlation-aware contrastive learning without masked video modeling (i.e. C-SCL). (c) and (f) correspond to MVCC(+C-SCL).
    }
    \label{tsnegradcam}
\end{figure}

\noindent
\textbf{Qualitative Analysis.} We further employed T-SNE and Gradient-weighted Class Activation Mapping (Grad-CAM) to visualize the representation distribution and model decisions. As shown in Fig.~\ref{tsnegradcam}, we mainly compared baseline, C-SCL and our complete framework. For the representation distribution, (b) and (c) is clearly better than (a) that the samples are more aggregated. Our method can make it easier for classifiers to construct decision boundaries. It can be observed from Grad-CAM visualization that the interpretability of high attention regions is poor on the baseline while the C-SCL model only focused on the area around the nodule. Our model precisely focused on the nodule regions during classification despite no location information being given.

\section{Conclusions}
In this paper, we proposed the first research about breast ultrasound video diagnosis for breast cancer. Thanks to masked video modeling and correlation-aware contrastive learning, the proposed learning framework can efficiently solve the challenge of video recognition under scarce supervised signals and complex nodule patterns. Experiments on a large-scale breast ultrasound video dataset prove the efficacy and flexibility of the proposed framework. Future research will focus on extending this framework to ultrasound videos of other organs and videos of other modalities.


\bibliographystyle{splncs04}
\bibliography{ref}

\end{document}